\documentclass[conference]{IEEEtran}
\IEEEoverridecommandlockouts
\usepackage{cite}
\usepackage{amsmath,amssymb,amsfonts}
\usepackage{graphicx}
\usepackage{textcomp}
\usepackage{xcolor}
\usepackage{algpseudocode}
\usepackage{algorithm}

\def\BibTeX{{\rm B\kern-.05em{\sc i\kern-.025em b}\kern-.08em
    T\kern-.1667em\lower.7ex\hbox{E}\kern-.125emX}}
\begin{document}

\title{Evolutionary Algorithms Approach For Search Based On Semantic Document Similarity.\\
}

\author{
\begin{tabular}[t]{c@{\extracolsep{8em}}c} 
Chandrashekar Muniyappa  & Eujin Kim \\
School of EECS & School of EECS \\ 
College of Engineering and Mines & College of Engineering and Mines \\
University of North Dakota & University of North Dakota \\
Grand Forks, ND 58202-7165 & Grand Forks, ND 58202-7165 \\
c.muniyappa@und.edu & ejkim@und.edu
\end{tabular}
}
\maketitle

\begin{abstract}
Advancements in cloud computing and distributed computing have fostered research activities in Computer science.  As a result, researchers have made significant progress  in Neural Networks, Evolutionary Computing Algorithms like Genetic, and Differential evolution algorithms. These algorithms are used to develop clustering, recommendation, and question-and-answering systems using various text representation and similarity measurement techniques. In this research paper, Universal Sentence Encoder (USE) is used to capture the semantic similarity of text; And the transfer learning technique is used to apply Genetic Algorithm (GA) and Differential Evolution (DE) algorithms to search and retrieve relevant top $N$ documents based on user query. The proposed approach is applied to the Stanford Question and Answer (SQuAD) Dataset to identify a user query. Finally, through experiments, we prove that text documents can be efficiently represented as sentence embedding vectors using USE to capture the semantic similarity, and by comparing the results of the Manhattan Distance, GA, and DE algorithms we prove that the evolutionary algorithms are good at finding the top $N$ results than the traditional ranking approach.
\end{abstract}


\section{Introduction}
Evolutionary computing refers to applying optimization techniques using computer programs known as evolutionary algorithms that are similar to the biological evolution process.  They are good at searching large solution spaces in a dynamically changing environment which is often difficult to code using deterministic rules, like determining the shortest path for robot navigation, simulating biological experiments, and many other optimization problems. There are many evolutionary algorithms, Genetic and Differential evolutionary algorithms are the most widely used algorithms. This research paper will show how they are applied to sentence embeddings generated by Universal Sentence Encoder to retrieve answers for the given user question. An overview of the genetic and differential evolution algorithms is shown in ``Algo. 1'' and ``Algo. 2'' respectively. Traditionally, the ranking approach is used to score the answers for the given question based on the similarity of the text and pick the top answers. However, research has shown that this approach may not always yield the best possible solution, especially when we want to retrieve the Top $N$ results. Even though, the first one or two answers match the question the rest of the answers may not be accurate. To overcome this problem we will apply evolutionary algorithms with advanced text embedding techniques to improve the overall quality of the Top $N$ search results.

\begin{algorithm}
\caption{Overview of the Genetic algorithm}
\begin{algorithmic}[1]
\State $Parents \gets \textit{\{randomly generated popualtion\}}$
\While{$\textit{not(termination criterion)}$}
    \State $\textit{Calculate the fitness of each parent in the population}$
    \State $Children \gets \emptyset$
    \While{$|Children| < |Paremrs|$}
        \State $\textit{Use fitness to Probabilistically}$
                $\textit{select a pair of parents mating}$
        \State  $\textit{Mate the parents to create the children $c_1$ and $c_2$}$
    \EndWhile
    \State $\textit{Randomly mutate some of the children}$
    \State $\textit{Parents} \gets \textit{Children}$    
\EndWhile

 \State \textit{Next generation}
\end{algorithmic}
\end{algorithm}
\begin{algorithm}
\caption{Overview of the Differential Evolution algorithm}
\begin{algorithmic}[1]
    \State $\textit{Set the generation counter, t = 0}$
    \State $\textit{Initialize the control parameters}$$\beta$, $p_r$
\State $\textit{Create and initialize the population, C(0), of n, individuals}$
\While{$\textit{stopping condition(s) not true}$}
     \For {$\textit{ each individual, x(t) C(t)}$}
        \State $\textit{Evaluate the fitness}$$f(x_i(t))$
    \State $\textit{Create the trial vector }$$U_i(t)$
    $\textit{,by applying the mutation operator}$
    \State $\textit{Create an offspring}$$x_i^{\prime}(t)$
    $\textit{by applying the crossover operator}$    
    \State $\textit{if}$ $f( (x_i^{\prime}))$ $\textit{better than}$ $f(x_i(t))$
    \State $\textit{Add}$ $(x_i^{\prime})$ $\textit{to}$ $C(t+1)$
    \State $\textit{Else}$
    \State $\textit{Add}$ $(x_i(t)$ $\textit{to}$ $C(t+1)$
    \State $\textit{end}$
    \EndFor
\EndWhile

\textit{Return the individual with the best fitness as the solution}
\end{algorithmic}
\end{algorithm}

The rest of the paper is organized into the following sections, the background is presented in section II, the approach in section III, the experiments and results in section IV, and finally, the conclusion and future work are presented in section V.

\section{Background}
Measuring text similarity is an important and challenging task in Natural Language Processing(NLP) that can be applied to retrieve documents matching users' requests. Jiapeng wang et al [1] in their study, described different types of text similarity measurements and the best text representation technique to be used. The study found that the string-based similarity measurement has the lowest performance as it can capture only lexical similarity and not semantic similarity, vector-based techniques make use of statistical methods to measure the semantic similarity between texts, and graph representation is used to identify the meaning of the same word in different contexts. However, the study concludes that the performance is relative to specific tasks and there is a need for the development of advanced algorithms. Jiyeon Kim et al [29] utilized semantic similarity to determine the similarity between music and movie two different content types to recommend relevant, but diverse content based on users' preferences. In addition, D. Meenakshi et al [30] developed the shared input LSTM (Long Short Term Memory) neural network to determine if a given set of documents are duplicates based on their semantic similarity, Manhattan distance was used to measure the similarity between documents. The work done by researchers in [29,30] shows how important it is to consider semantic similarity over string-based similarity. In this research, we will represent text as embedding vectors generated by the Universal Sentence Encoders (USE) [2]. USE is a deep neural network (DNN) model that is pre-trained on millions of documents, it can be integrated with other machine learning models using the transfer learning approach [5]. Therefore, researchers can readily use the model to encode the text into vectors without the need for any training data which is usually a challenging step in NLP tasks. Besides, the USE is based on an advanced general-purpose encoding scheme that effectively captures the semantic similarity of texts of variable lengths. The results presented in USE [2] clearly show improved performance when compared to other techniques. S. Velampalli et al [3] applied USE on Twitter comments to build a sentiment classification model using the transfer learning technique. In the research, they show how transfer learning can be effectively used to integrate different Machine Learning models to increase the speed of development and also achieve overall high accuracy. 

\par
Ranking documents to retrieve relevant documents based on user search is an old problem [7] many researchers have studied this area for a long time and have devised many algorithms and metrics to evaluate the performance of the ranking system. Ranking documents for search results is a challenging task, as the data may keep changing, and measuring the similarity based on the content is difficult. Besides, it is often challenging to factor in all these changes as part of the ranking fit function. Considering all these challenges, D. Bollegala et al [6] invented a novel RankDE algorithm based on the Differential Evolution algorithm to retrieve the most relevant documents based on user search. Similarly, Urszula Boryczka et al [8], applied the Differential Evolution (DE) algorithm [9]  to create a personalized list of recommended items. As part of this research, they used Singular Value Decomposition (SVD) [10], the matrix factorization technique to generate the real numbers based on the users-to-items rating relationship for each user and represented them as the chromosomes to build a recommendation system. They make use of Mean Average Precision (MAP) shown in  equation \eqref{ap2}, as the fitness function, to measure the quality of the recommended items.

\begin{equation}\label{ap1}
P@n=\frac{\textit{Number of relevant items in top n results}}{n}
\end{equation}
Average precision (AP) averages the $P@n$ for different n values:
\begin{equation}\label{ap2}
AP=\frac{\sum_{n=1}^{N}(P@n \times rel(n))}{\textit{Number of relevant items for this query}}
\end{equation}
Where
\begin{itemize}
\item ${rel(n)}$ is a binary function assigning the value 1 if the nth document is relevant to the query (otherwise 0) and N is the number of items obtained. Mean Average Precision (MAP) averages the AP values for all ${U}$ users in the system 
\end{itemize}
\par
P. Sihombing et al [4] applied the Genetic algorithm to search and retrieve the documents based on users’ inputs. In this study, keywords  (String-based similarity measurement) in the document were converted into a binary sequence to represent the chromosomes, and Dice distance measurement was used to measure the fitness score. 
B. Alhijawi et al [11], applied the Genetic algorithm to build a collaborative recommendation system. They represented the similarity between users and items as vectors of randomly generated floating point values and 
the fitness was measured using the Mean Absolute Error (MAE) between the predicted and known user ratings. The lesser the value of MAE better is the results. 
\par
Alan Diaz-Manriquez et al [12] used the graph-based representation to cluster the documents based on ACM taxonomy using the Genetic algorithm. Even though the clustering was based on taxonomy, the keywords in the documents were used to represent the chromosomes, and similarity was measured using the Floyd-Warshall algorithm [13]. To measure the cluster purity Davies-Bouldin index (DB Index) [14] was used, which measures the radius of clusters, the smaller the value better is the performance. To represent the chromosomes, keywords in the documents were assigned unique numbers to generate the floating-point vectors. In order to identify the context of words, a graph was used where each node represents different categories. In this way, they can differentiate the meaning of the same words in different contexts. K. Nandhini et al [15] applied the Differential evolution algorithm to extract summaries from texts. In the research, firstly, they cleaned up text using NLP steps like tokenizing, stemming, and removing stopwords. Secondly, they applied the Mathematical Regression model [16] to generate the text embeddings. Finally, Cosine similarity [17] was used to build a similarity matrix for the given set of sentences. The sentences were numbered sequentially, and these numbers were used to represent the chromosomes. The fitness function was used to find combinations of sentences that produce maximum information score and cohesion. As the chromosomes represented the sequence of sentences, the default implementation of mutation and crossover would not retain the sequence. Hence, both operators were supplied with a custom implementation that will ensure sentence ordering is maintained throughout the mutation process. 
\par
R.J. Kuo et al [25] applied the Differential Evolution algorithm along with Restricted Boltzmann Machine (RBM) and K-means algorithm to cluster users and build a recommendation system. RBM is widely used in deep learning, however, tuning the parameters is challenging, as part of this research, they applied DE to tune the RBM algorithm's parameters namely Learning, Momentum, and Weight decay. Finally, they used the model to build a recommendation system and through experiments proved that this approach is better than GA and Particle Swarm Optimization (PSO) recommendation systems. U. Boryczka et al [26] applied the Differential Evolution algorithm to rank and select the top $N$ recommendation items, to improve the efficiency without reducing the quality of the top $N$ items selected they applied dot product to score the user profiles. The user-to-item relationship was modeled as a matrix and the population generated by DE was also converted to a matrix. Lastly, the dot product was computed between the matrices to score and pick the top $N$ similar documents. Through experiments, they proved that the similarity computed by applying DE with dot product produced similar results when compared to standard similarity measurement algorithms like Cosine similarity, Euclidean distance, Manhattan distance, and so on, but with a significant gain in the run-time. 
\par
D.Mustafi et al [27] applied a combination of the Genetic and Differential Evolution Algorithm, a hybrid approach to cluster text documents using the K-means clustering algorithm. The traditional K-means algorithm [28] has a local optimization problem, therefore, it usually returns lesser than the requested number of clusters by assigning all the data points to fewer clusters. To overcome this problem, researchers proposed a novel centroids selection algorithm; instead of randomly selecting them. As part of this approach, instead of picking the centroids randomly, they pick a centroid that is central to the entire dataset (all the clusters), and subsequent centroids are picked, such that, they are uniformly distributed across the dataset. In this way, the centroids with different features are selected to make sure the requested number of clusters are built instead of returning empty clusters. To represent chromosomes, they encode documents using Term-Frequency (TF) and Inverse-Document-Frequency (IDF) [29] values of each keyword. And, Cosine similarity was used with both GA and DE algorithms to measure the similarity between documents and clusters. Through experimental results, they proved that the results obtained by the hybrid algorithm were better than the traditional K-means algorithm.

\section{Approach}
\subsection{ Text Representation}
\begin{figure}
    \centering
    \includegraphics[width=8cm, height=4cm]{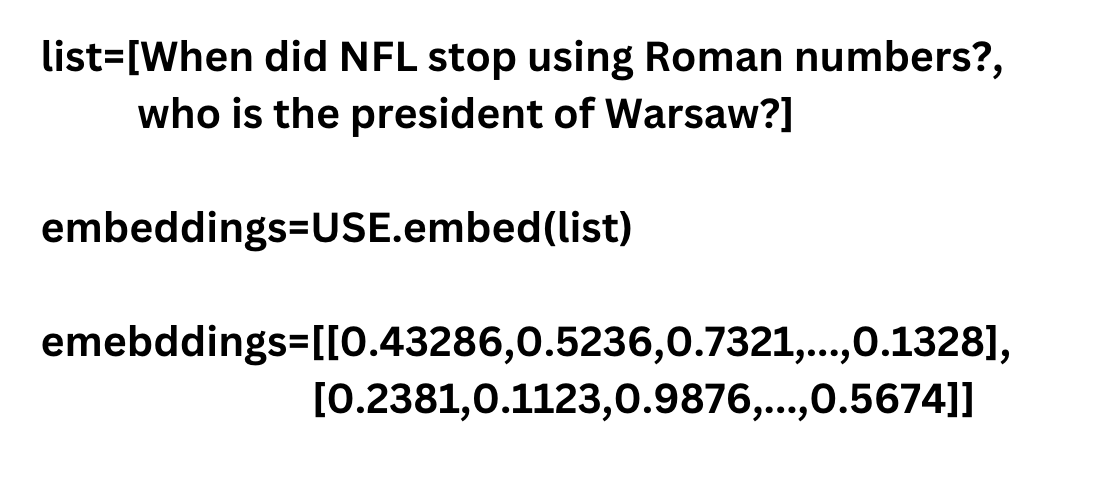}
    \caption{USE Embeddings}
    \label{fig:use_embeddings}
\end{figure}

Based on the literature survey, one can observe that researchers used different types of text representation techniques like binary, vector embeddings, and graph embeddings to represent chromosomes and applied variations of Genetic Algorithm (GA) and Differential Evolution (DE) algorithms to measure text similarity. However to generate the vector embeddings, words were assigned unique integers or random floating point values. The main problem with these approaches is different words with the same meaning will get different values without taking synonym property into consideration, as a result, semantic meaning will not be captured despite using the adaptive evolutionary algorithms. To overcome this problem, we will use embeddings generated by Universal Sentence Encoder (USE) which will capture the semantic similarity between the words and generate fixed-length embeddings. As shown in Figure \ref{fig:use_embeddings}, text can be directly fed into the USE model without applying any NLP cleaning steps [18].

The model will take a list of sentences, and paragraphs of variable length and generate fixed-length embeddings of length 512 for each element in the input list. The model uses a general-purpose encoding scheme mainly designed for transfer learning purposes. For this research, we are using the "Stanford Question Answering Dataset" [19] which contains more than 100 thousand questions and answers. We will encode all the questions using USE and use it as the population for GA and DE algorithms.
\subsection{ Chromosomes}
\begin{figure}[h]
    \centering
    \includegraphics[width=8cm, height=4cm]{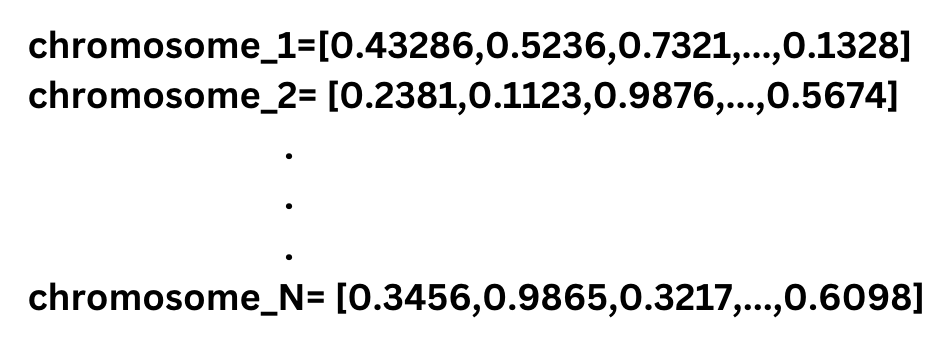}
    \caption{Chromosomes}
    \label{fig:chromosomes2}
\end{figure}

 Chromosomes play a pivotal role in evolutionary algorithms. There are different ways of representing the chromosomes like integer, binary, and floating point values. However, research has proved that floating point representation has the best performance [20]. Therefore, word embeddings generated as described in the previous section will be used to represent the chromosomes each of length 512 containing floating point values as shown in Figure \ref{fig:chromosomes2}.

\subsection{ Fitness function}
The fitness function is used to guide algorithms to search large solution spaces. The Manhattan distance, shown in equation \eqref{manhattan} was used as the fitness function to measure the similarity of text documents.

\begin{equation}\label{manhattan}
Sim= mean(|\Vec{X_1} - \Vec{X_2}|)
\end{equation}

Where
\begin{itemize}
\item ${\Vec{X_1}}$ is the first text document embedding vector of length 512.
\item ${\Vec{X_2}}$ is the second text document embedding vector of length 512.
\end{itemize}

In this research, $\Vec{X_1}$ will be the question embedding input by a user, and $\Vec{X_2}$ will be the embedding of different questions from the dataset. The algorithm will measure the similarity of the user input question with all the questions in the dataset and select the question which has the lowest mean value, meaning, it will have the highest similarity. Therefore, the objective is to minimize the fitness function value.
\subsection{ Genetic Algorithm approach}
To implement Genetic Algorithm (GA) [22], PyGAD (Python Genetic Algorithm) open-source library was used [21]. The algorithm was set up with the following configurations. Firstly, the population was initialized to all the question embeddings generated using USE, and $100$ parents were selected in each iteration for mating (mating pool). Secondly, the "Steady-State" parent selection type was used to select parents for mating, and the three best solutions were propagated to the next generation as part of the elitism concept. Finally, the "Single-Point" Crossover with "Random" Mutation operators was used.
\subsection{ Differential Evolution Algorithm approach}
The Differential Evolution algorithm [23] was implemented using the TensorFlow Probability (TFP) open-source library [24]. The "DE/rand/1/bin" scheme was used in this research. The population was initialized to the question embeddings generated using USE and parents were selected randomly for mutation. The scaling factor $\beta$ was set to 0.5 which controls the magnitude of mutation. In addition, one difference vector with a binomial cross-over operator was used with a cross-over probability of 0.9.     
\par

The steps followed to apply GA and DE algorithms are shown in "Algo. 3".

\begin{algorithm}
    \caption{GA and DE Steps}
    \begin{algorithmic}[1]
        \State \textit{Q is the Set of all the questions}
        \State \textit{Generate the embedding vectors $\Vec{Q_e}$ for the set Q by applying USE}
        \State \textit{Initialize the population to $\Vec{Q_e}$}
        \State \textit{A user will submit the question q}
        \State \textit{Generate the embedding vector $\Vec{q_e}$ for q}
        \State \textit{Apply the GA and DE algorithms with the objective function, equation \eqref{manhattan} to measure the similarity between $\Vec{q_e}$ and $\Vec{Q_e}$ embedding vectors and select the Top $N$ answers.}
    \end{algorithmic}
\end{algorithm}
\section{Experiments And Results}
The GA and DE algorithms were applied to the SQuAD dataset [19] independently as described in "Algo. 3" with different configurations. Finally, the optimal performance was observed with the configurations described in their respective sections. Besides, the Manhattan distance was applied to measure the similarity without GA, and DE algorithms, and the results of all three algorithms are shown in Figures [3-5]. The equation \eqref{manhattan} was used as the objective function for GA and DE algorithms. As one can see, the results of Manhattan distance, GA, and DE algorithms shown in Figure 3, Figure 4, and Figure 5 respectively are similar, with accurate results at the top and inaccurate matches at the bottom of the list. This approach works for the Top 1 or 2 answers, but not for the top 10 answers. To solve this problem, instead of considering only the optimal answer returned by the objective function. Let us also consider the top 2 suboptimal answers of the GA and DE algorithms from various generations.

\begin{figure}[ht]
    \centering
    \includegraphics[width=9cm, height=7cm]{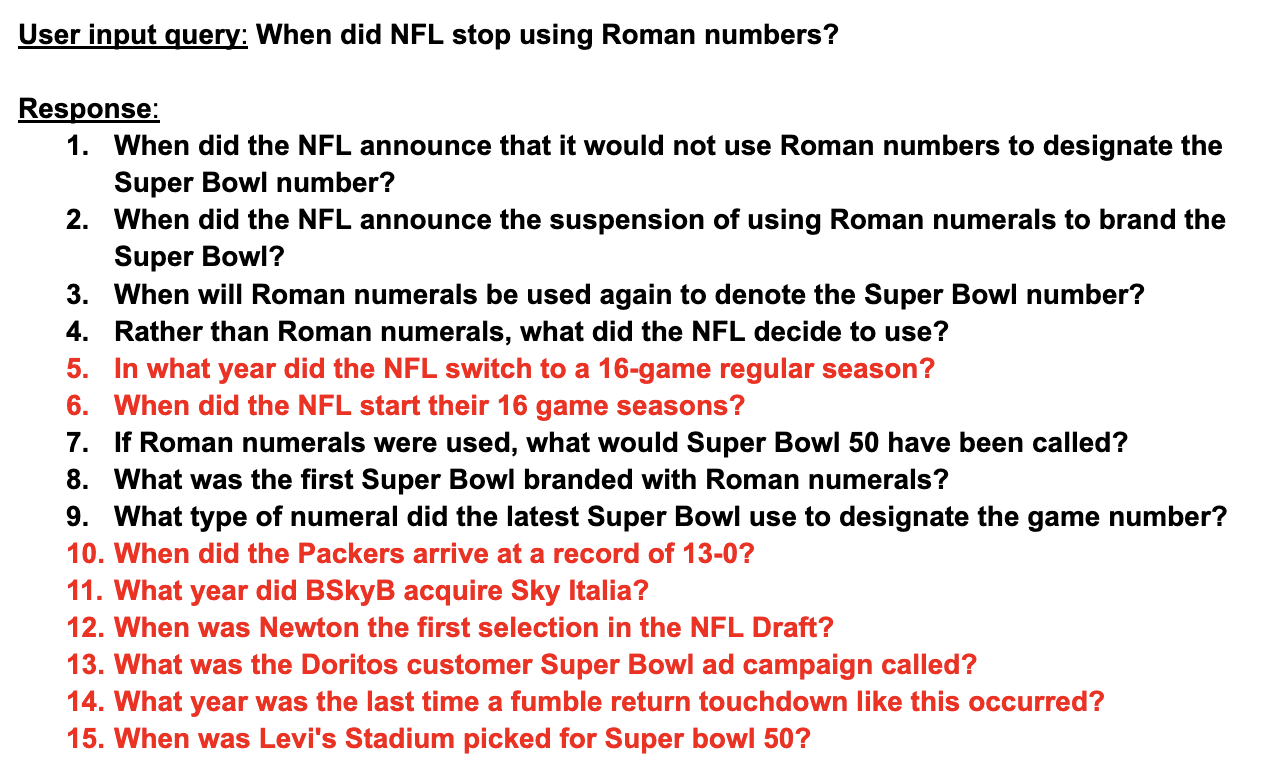}
    \caption{Manhattan Similarity Results}
    \label{fig:chromosomes1}
\end{figure}

\begin{figure}[h!]
    \centering
    \includegraphics[width=9cm, height=7cm]{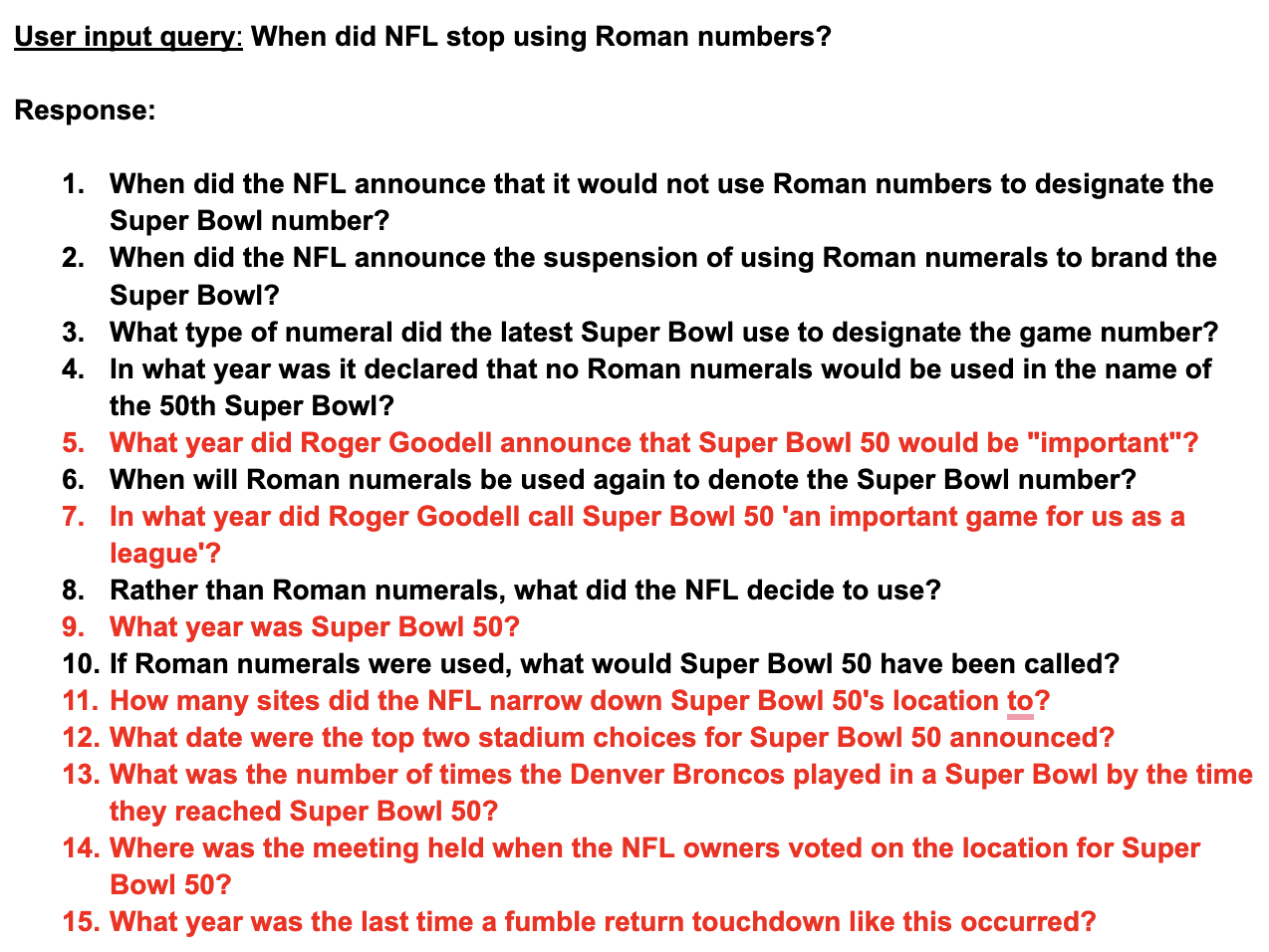}
    \caption{GA Optimal Result}
    \label{fig:chromosomes}
\end{figure}

\begin{figure}[h!]
    \centering
    \includegraphics[width=9cm, height=7cm]{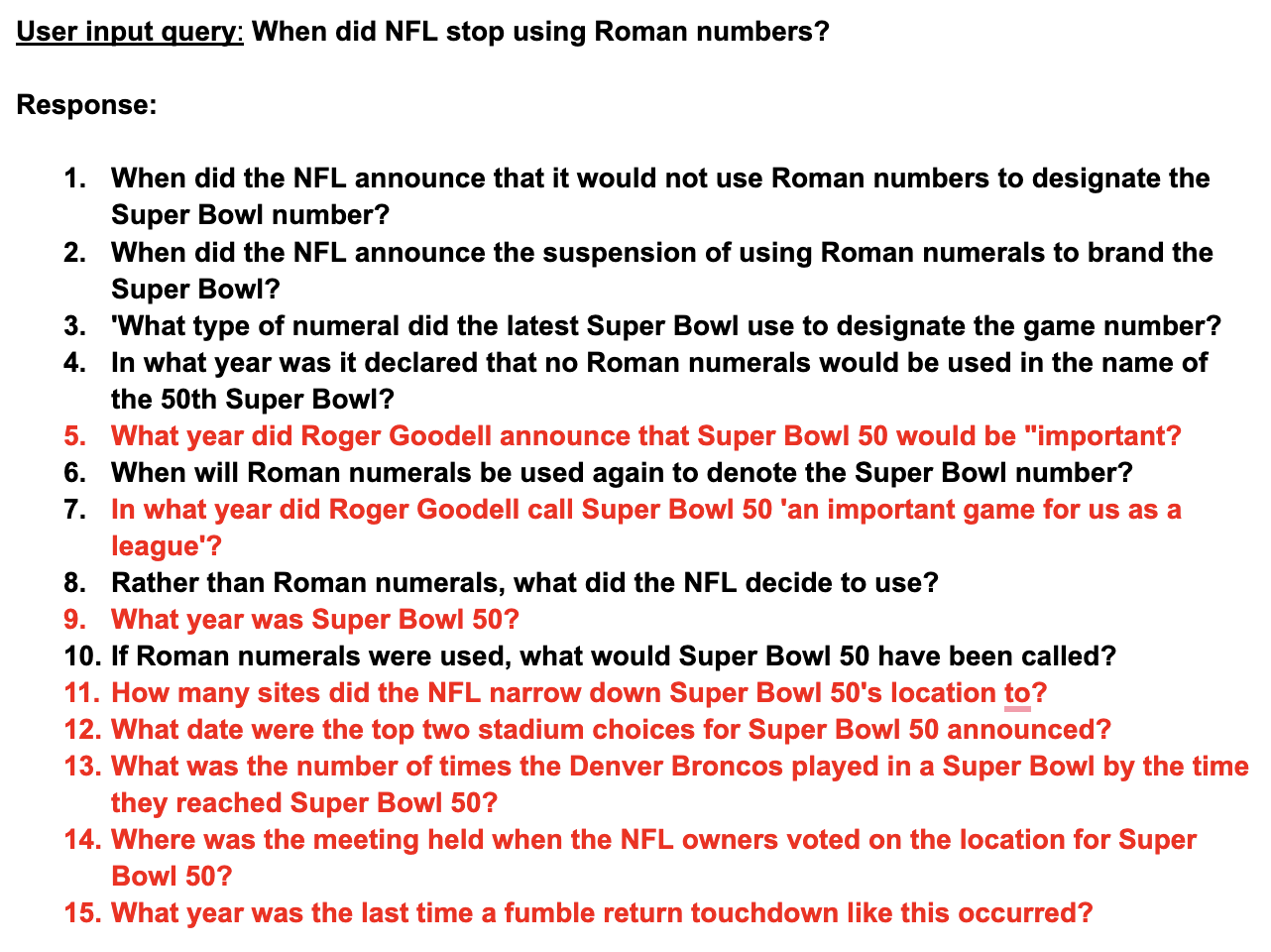}
    \caption{DE Optimal Result}
    \label{fig:chromosomes3}
\end{figure}

\begin{figure}[h!]
    \centering
    \includegraphics[width=9cm, height=7cm]{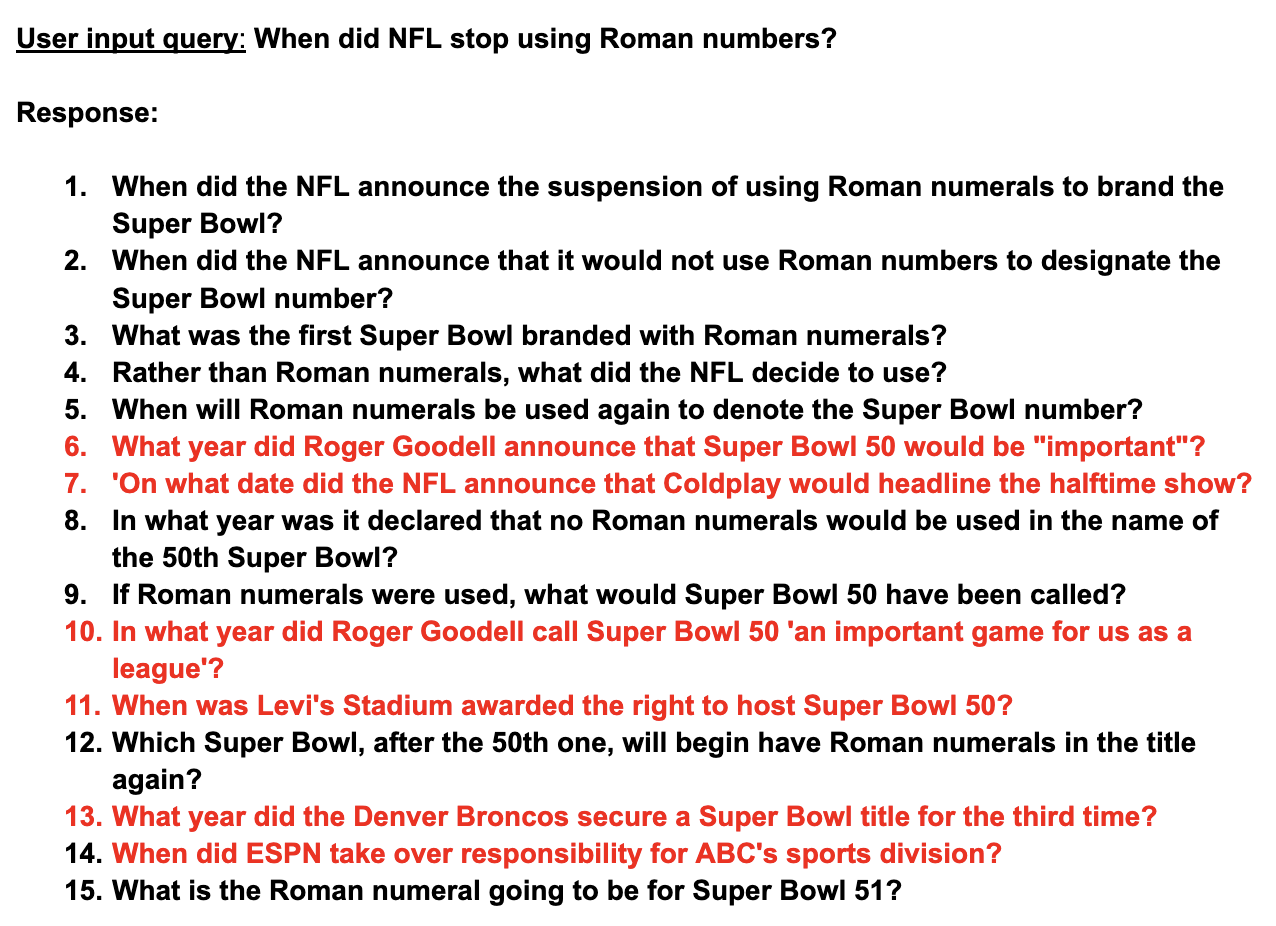}
    \caption{GA Suboptimal Result: 1}
    \label{fig:chromosomes4}
\end{figure}

\begin{figure}[h!]
    \centering
    \includegraphics[width=9cm, height=7cm]{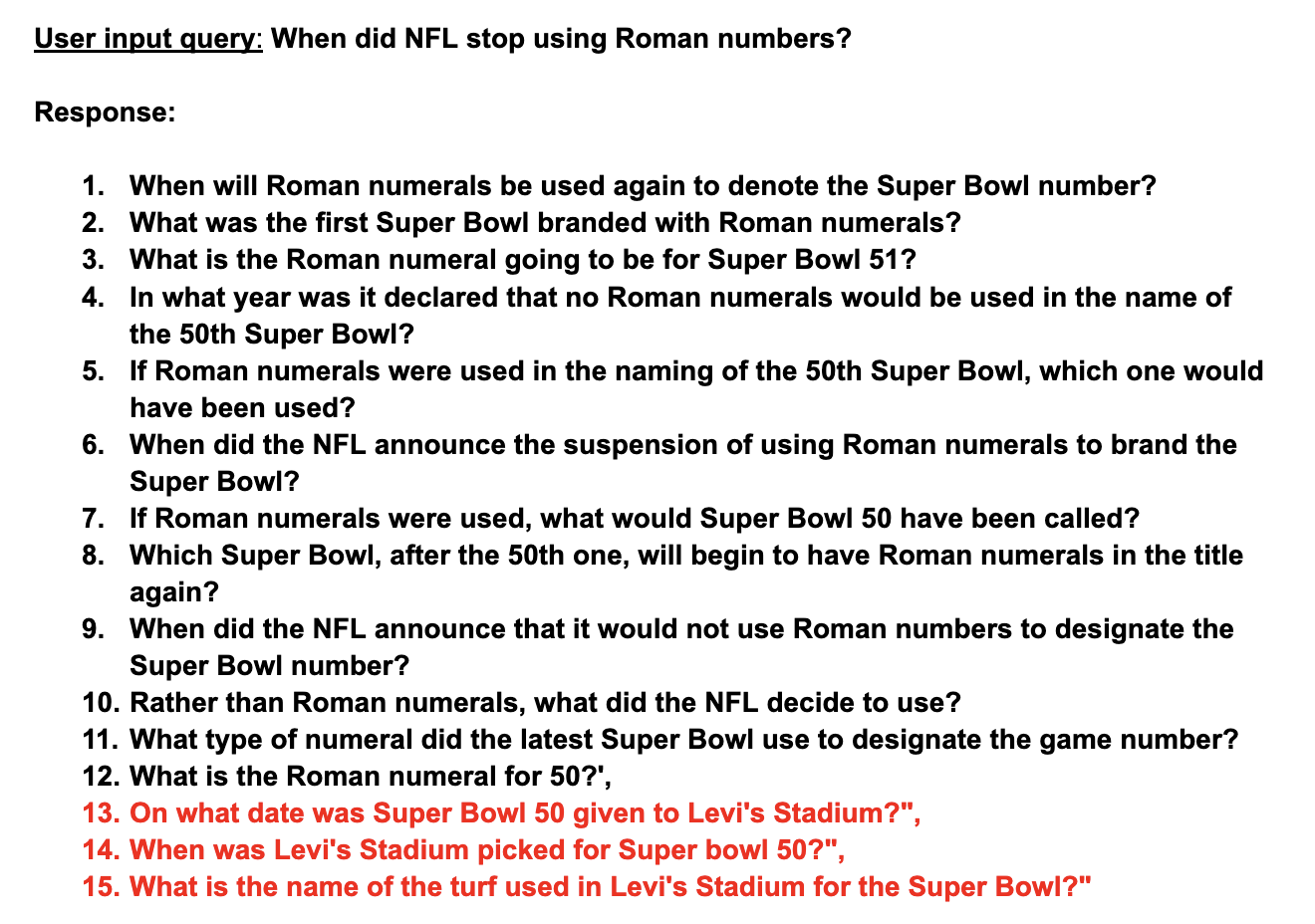}
    \caption{GA Suboptimal Result: 2}
    \label{fig:chromosomes5}
\end{figure}

\begin{figure}[h!]
    \centering
    \includegraphics[width=9cm, height=7cm]{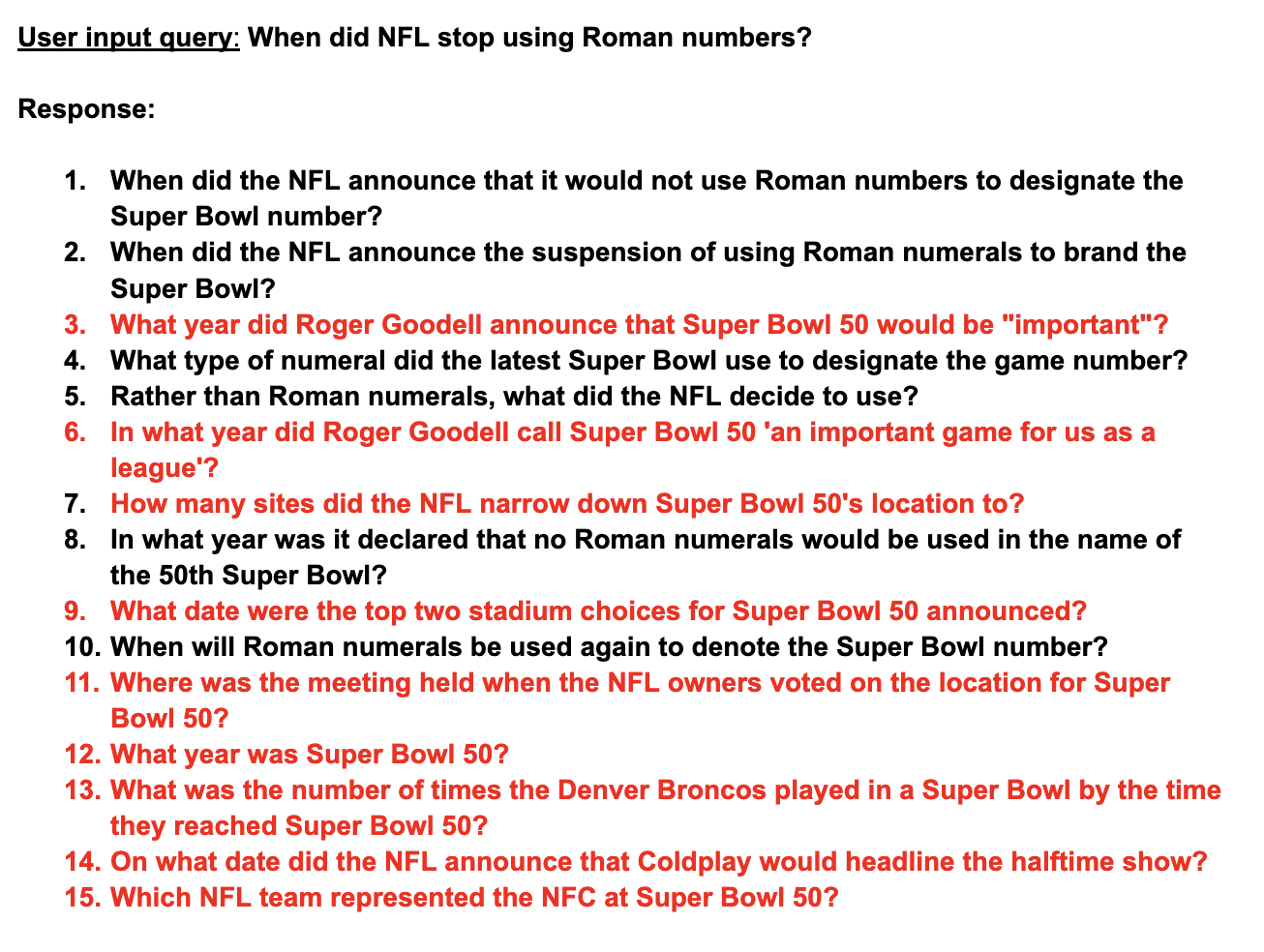}
    \caption{DE Suboptimal Result: 1}
    \label{fig:chromosomes6}
\end{figure}

\begin{figure}[h!]
    \centering
    \includegraphics[width=9cm, height=7cm]{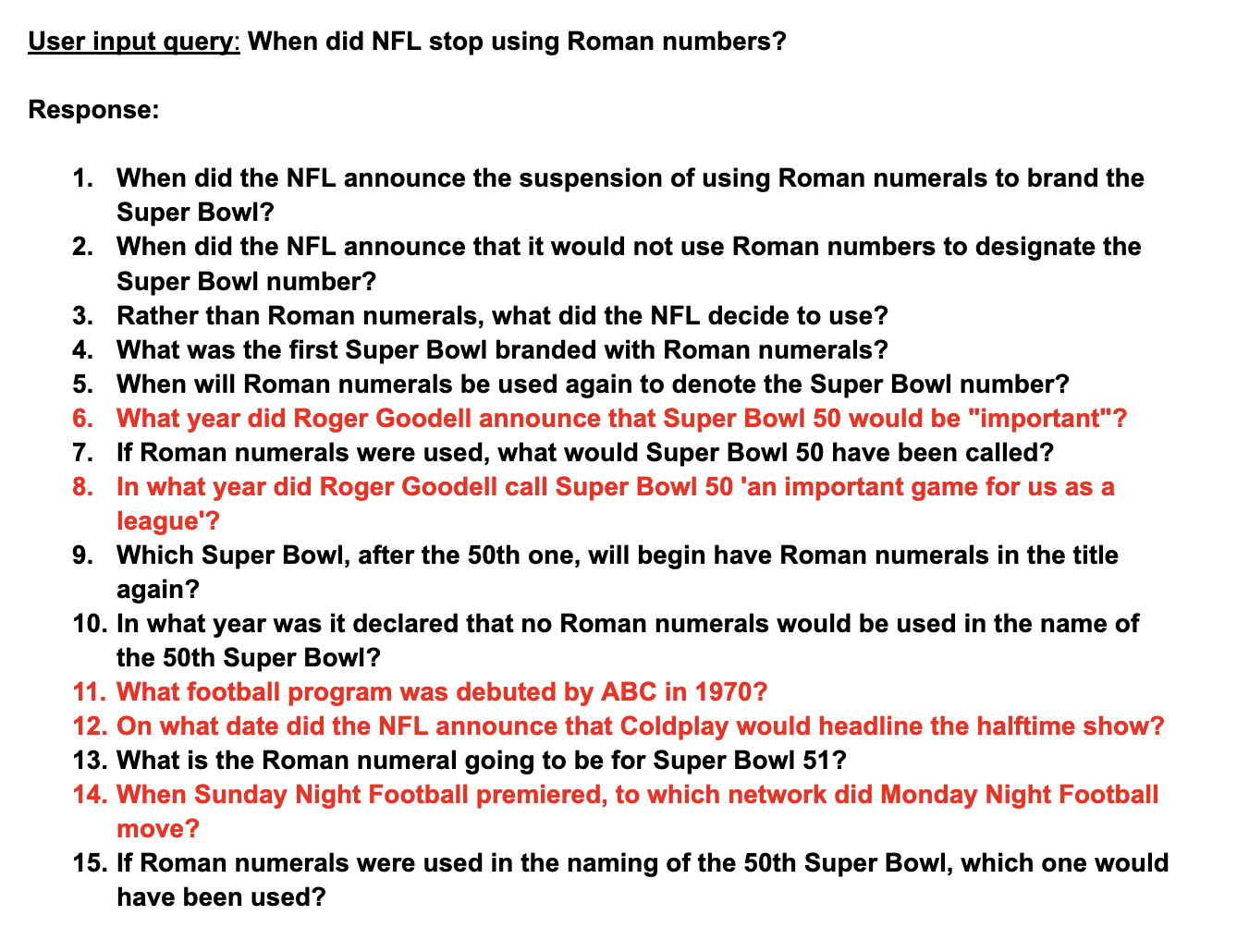}
    \caption{DE Suboptimal Result: 2}
    \label{fig:chromosomes7}
\end{figure}

When we observe Figure 4, Figure 6, and Figure 7 optimal result, and suboptimal results 1 and 2 of the GA algorithm respectively. We can see that, the optimal result in Figure 4 has an exact match at the top and noise at the bottom of the list. However, the suboptimal results in Figure 6 and Figure 7 have more relevant answers all the way to the bottom of the list, but the most accurate answers at the top of the list are dropped. On the contrary, when we compare the DE algorithm results in Figure 5, Figure 8, and Figure 9, we can observe that the exact match answer at the top of the list is retained in both optimal and suboptimal results. In addition, noise at the bottom of the list is also reduced. This clearly demonstrates the power of the evolutionary algorithms over the traditional ranking approach to fetch relevant Top N documents. Besides, we can also observe that many documents with semantic meaning are part of the result, demonstrating the power of sentence embedding and transfer learning techniques. In addition, when we apply human annotations 1 when an answer is relevant and 0 otherwise and apply the equation \eqref{ap2} it will result in Figure 7 and Figure 9 as the results for the GA and DE algorithms respectively, which will result in reduced accuracy of the final resultset. Furthermore, defining a single objective function to capture all the optimal answers in a document retrieval problem is challenging [6]; Therefore, it is important to consider the suboptimal results  along with the optimal result and post-process the resultsets to build the final Top N results.
\section{Conclusion and Future work}
In this research paper, we have applied the GA and DE evolutionary algorithms on sentence embeddings generated by USE, using the transfer learning technique to search and retrieve the documents based on semantic similarity. By applying the suggested approach to the "SQuAD" question and answer dataset, we show how text can be efficiently represented as sentence embeddings vectors to capture the semantic similarity using USE. In addition, by comparing the results of Manhattan distance, GA, and DE algorithms we prove that evolutionary algorithms are good at retrieving the Top N search results than the traditional ranking approach. However, due to the nature of the document search and  retrieval problem, it is difficult to design a single objective function to capture all the optimal results. Therefore, suboptimal results have to be considered and post-processing has to be done to build the final resultset. As part of future work, we can consider multi-objective evolutionary algorithms that can effectively capture and return the results without the need to examine and post-process the suboptimal results.

\end{document}